# Collaborative Attention Memory Network for Video Object Segmentation

Zhixing Huang[1], Junli Zha[†], Fei Xie, Yuwei Zheng, Yuandong Zhong, Jinpeng Tang

Alibaba Group, Guangzhou, China

## Abstract

*Semi-supervised video object segmentation is a fundamental yet Challenging task in computer vision. Embedding matching based CFBI series networks have achieved promising results by foreground-background integration approach. Despite its superior performance, these works exhibit distinct shortcomings, especially the false predictions caused by little appearance instances in first frame, even they could easily be recognized by previous frame. Moreover, they suffer from object's occlusion and error drifts. In order to overcome the shortcomings , we propose Collaborative Attention Memory Network with an enhanced segmentation head. We introduce a object context scheme that explicitly enhances the object information, which aims at only gathering the pixels that belong to the same category as a given pixel as its context. Additionally, a segmentation head with Feature Pyramid Attention(FPA) module is adopted to perform spatial pyramid attention structure on high-level output. Furthermore, we propose an ensemble network to combine STM network with all these new refined CFBI network. Finally, we evaluated our approach on the 2021 Youtube-VOS challenge where we obtain 6th place with an overall score of 83.5%.*

## 1. Introduction

Semi-supervised video Object Segmentation(VOS) aims at segmenting a particular set of objects in each frame of a video based upon the set of ground truth object masks given at the first frame. The VOS proved to be crucial in many industrial applications, including video editing, autonomous driving, etc.

Recently embedding matching based methods[1][6] make great advances in VOS. Space-Time Memory Network(STM)[5] introduces memory networks to learn to read sequence information, boosting the performance a lot. The past frames with object masks are formed as an external memory, which is used by the current query frame for segmenting. Moreover, the query and the memory are densely matched in the feature space, covering all space-time pixel locations in a feed-forward fashion. By using multi-frame information, it achieved higher accuracy in many cases, including appearance changes, occlusions, and drifts. By considering foreground-background integration, CFBI series networks[8][7] achieve the new state of the art. CFBI series networks treat the background equally with the foreground, which extract the embedding and do match for both foreground target and background region. Moreover, pixel-level and instance-level embeddings are extracted for each video frame to improve the accuracy for different scales of features.Furthermore, a collaborative ensemble is used to aggregate all these four information and implicitly learn their relationship.

Even though the efforts mentioned above have made significant progress, these works not introduce the specific attention mechanism for object. Self-attention has proved to be an efficient method for image segmentation. In this paper, we propose a Collaborative Attention Memory Network, which follow CFBI and examine a collection of object attention module that affect the video object segmentation performance. First, object context from OCNet[9] is used to define the set of pixels that belong to the same category, emphasizing the object pixels that are essential for labeling the pixel. Second,Feature Pyramid Attention module from Pyramid Attention Network(PAN)[4] is adopted to perform spatial pyramid attention structure on high-level output. Moreover, STM is combined to improve the performance for occlusions. Furthermore, an ensemble network is designed to combine all these networks, which creates a significant improvement in the overall score. Finally, we introduce a new post-processing methods to remove the explicit bad cases, including track losing, little tiny object segmentation.

## 2. Method

Our method is derived from the CFBI+[7] and STM

---

[1] Equal contribution.

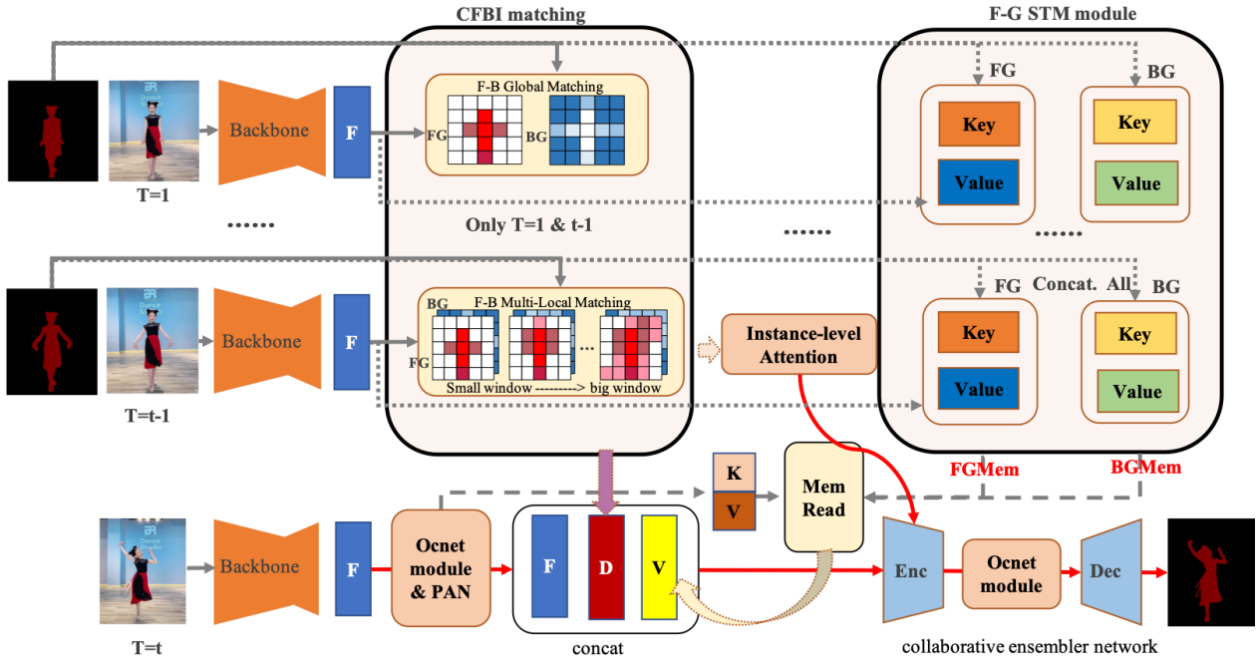

Figure 1: Framework of Collaborative Attention Memory Network. The CFBI matching only needs the first frame (t = 1) and the previous frame (t = T − 1), and the STM module needs all the past frames.

framework. The structure of Collaborative Attention Memory Network is illustrated in Figure 1. To be specific, we implant CFBI+ F-G pixel-level matching and instance-level attention in our framework. Furthermore, we apply the thought into STM module. Not only learning key and value embedding from foreground embedding, but also our framework considers embedding learning from background embedding for collaboration. The detail of F-G STM module is explained in Sec.2.1. We introduce the object context in Sec.2.2. And we use ensemble modeling which is described in Sec.2.3. Finally, we introduce the post-processing in Sec.2.4.

### 2.1. Foreground-Background Space-Time Memory Module

In order to reduce the number of parameters and the GPU memory, we firstly extract the all-frames embeddings by using the same backbone network. In the module, we separate embeddings into the foreground and background pixels based on their masks. It is a more worthy research topic on how to fuse their masks. Thus, we propose two different schemes to separate embeddings. At the very start, extract the embedding.

(1) As shown in Figure 2(a) , we separate mask into foreground and background, then the masks are multiplied with the embedding to produce foreground feature map and background feature map.

(2) The second schemes as shown in Figure 2(b), the mask is concatenated with current frame embedding to get a feature map, then we use a convolution layer with kernel size 1x1 and the sigmoid layer to produce a spatial prior. The prior is multiplied with the embedding.

Finally, echo feature map through two parallel convolutional layers, outputs two feature maps – key and value.

### 2.2. Object Context

We apply OCNet module in our frameworks, which focuses on enhancing the role of object information. In order to reduce the computational complexity, we only apply the approach on the feature map with output stride 16, and also demonstrate the effectiveness of the module.

PAN firstly accomplished bottom-up path augmentation which follow FPN[2] to define that layers producing feature maps with the same spatial size are in the same network stage.

By fusing feature maps from different spatial size, information diversity increases and masks with better quality are produced, especially the effect of small-scale object segmentation. So, we apply the module in our backbone stage.

## 2.3. Ensemble Modeling

Ensemble modeling is a process where multiple diverse base models are used to predict an outcome. We have trained many different models, but they alone could not achieve a very high precision. Through the analysis of the resultswe find that these models are complementary. Thus, we believe the ensemble modeling can effectively improve the performance. Many styles of ensemble modeling exist today, including voting, average, stacking, etc. In this paper, we trained a light network to ensemble modeling as shown in Figure 3, which achieves better segment performance.

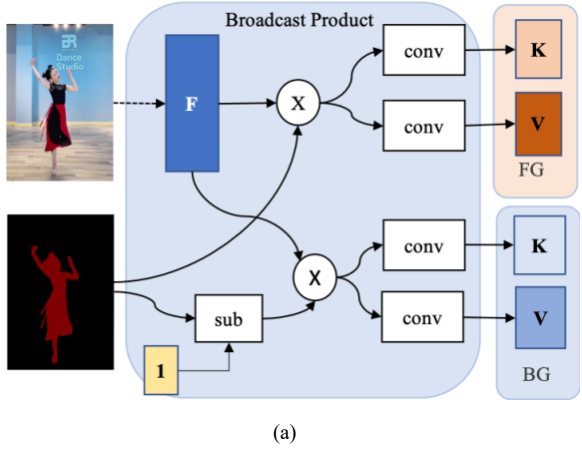

(a)

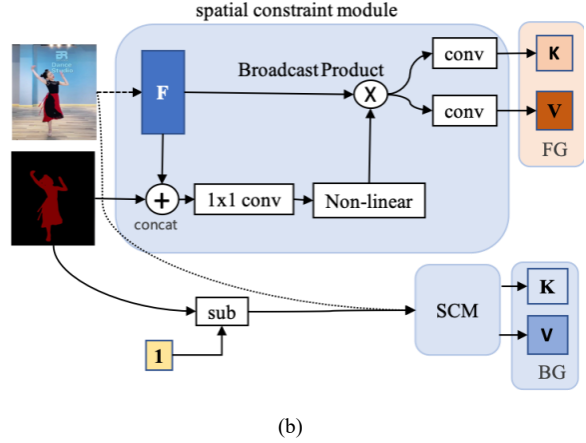

(b)

Figure 2: The F-G STM module

our framework for fine segmentation.The Result is shown in Figure 4.

## 3. Experiments

In the challenge, our method only uses ResNet-101 as the backbone which had been pretrained on COCO[3], and we did not use additional data sets for training. For the training, we use a learning rate of 0.01 with a batch size of 8 using Tesla V100 32G GPUs. And we also use sequential training method. For the testing, the multi-scale input size of the network for inference is adopted to boost the performance.

### 3.1. Results

Our proposed method achieves overall score of 83.7% on the YouTube-VOS 2019 validation set, and we achieve overall score of 83.5% in the YouTube-VOS 2021 testchallenge as shown in Table 1. So, our scheme has strong robustness, we obtained relatively high performance in both the validation set and the test set.

### 3.2. Ablation Study

The ablation study of the proposed components is shown Table 2. The baseline is our re-implementation of CFBI which achieves 81.3% on YouTube-VOS 2019 validation set, and the performance improve from 81.3% to 83.7% when we adopt all components.

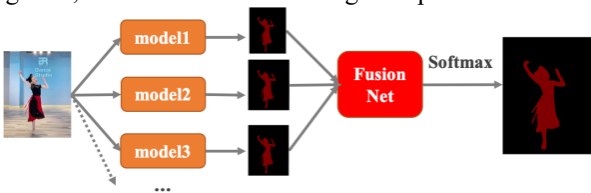

Figure 3: ensemble modeling.

## 2.4. Post-processing

We propose two effective post-processing methods, which can remove the obvious errors and improve the performance of small object segmentation. First, based on the principle that the object movement between adjacent frames is unlikely to be very large. When the connected domain of the mask is detected, the object mask close to the last position is retained, and the one too far away is removed directly. Second, according to the mask of the first segmentation, the area of the small object is calculated, then the original image is cropped and we predict mask again by

## 4. Conclusion

In this work, we have presented a novel Collaborative Attention Memory Network for the semi-supervised video object segmentation. Our approach achieves an Overall of 83.5% on the Youtube-VOS 2021 semi-superviesd video object segmentation challenge.

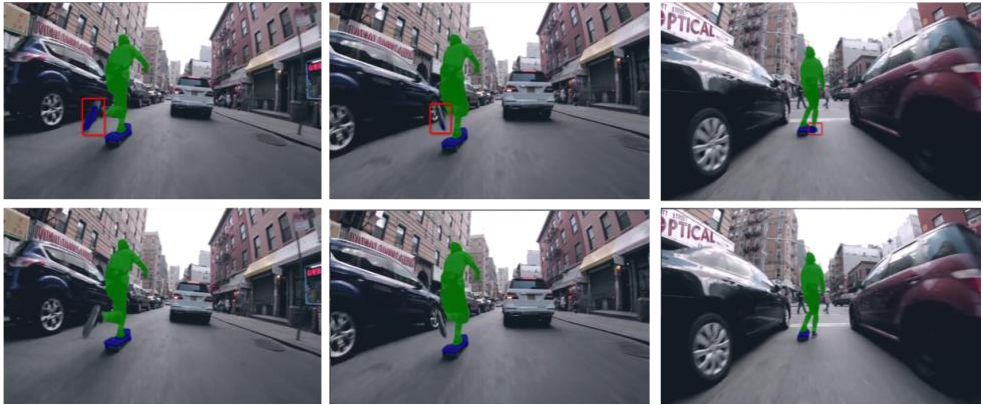

Figure 4: Qualitative comparison between origin and post-processing. The first row shows the origin predictions, and the red bounding box is the error predictions. The second row shows the correct predictions after post-processing.

| Team | Overall | J$_{seen}$ | J$_{unseen}$ | F$_{seen}$ | F$_{unseen}$ |
|---|---|---|---|---|---|
| wenhaowang | 0.856 | 0.836 | 0.811 | 0.888 | 0.889 |
| hkchengrex | 0.854 | 0.828 | 0.814 | 0.883 | 0.893 |
| qinghualiuyong | 0.842 | 0.816 | 0.799 | 0.870 | 0.881 |
| cncyww | 0.839 | 0.823 | 0.788 | 0.874 | 0.871 |
| cheng321284 | 0.836 | 0.809 | 0.798 | 0.859 | 0.877 |
| PixelKitty(**ours**) | 0.835 | 0.814 | 0.793 | 0.866 | 0.868 |

Table 1: Ranking results in the YouTube-VOS 2021 test-challenge.

| Baseline and components | Overall | ∆ |
|---|---|---|
| Baseline (CFBI re-implementation) | 81.3 | - |
| +FPN | 81.6 | +0.3 |
| +PAN | 82.0 | +0.7 |
| +OCNet: | 82.1 | +0.8 |
| +Flip and multi-scale | 82.9 | +1.6 |
| +Post-processing | 81.5 | +0.2 |
| STM + ensemble modeling | 83.6 | +2.3 |
| Final (STM + ensemble modeling + post-processing) | 83.7 | +2.4 |

Table 2: The ablation study experiments on YouTube-VOS 2019 validation set.